\documentclass[sigconf]{acmart}

\AtBeginDocument{%
  }

\copyrightyear{2026}
\acmYear{2026}
\setcopyright{cc}
\setcctype{by}
\acmConference[MM '26]{Proceedings of the 34th ACM International Conference on Multimedia}{November 10--14, 2026}{Rio de Janeiro, Brazil}
\acmBooktitle{Proceedings of the 34th ACM International Conference on Multimedia (MM '26), November 10--14, 2026, Rio de Janeiro, Brazil}
\acmDOI{10.1145/3767308.3836282}
\acmISBN{979-8-4007-2213-4/2026/11}
\usepackage{multirow}

\begin{document}

\title{PWLR: Pairwise Witness Local Rejection for Boundary-Aware Out-of-Distribution Detection}

\author{Chengyao Jia}
\orcid{0009-0008-8653-4463}
\affiliation{%
  \department{Guanghua School of Stomatology}
  \institution{Sun Yat-sen University}
  \city{Guangzhou}
  \country{China}}
\affiliation{%
  \department{Guangdong Provincial Key Laboratory of Stomatology}
  \city{Guangzhou}
  \country{China}}
\email{jiachy5@mail2.sysu.edu.cn}

\author{Ruixuan Wang}
\orcid{0000-0002-8714-0369}
\correspondingauthor
\affiliation{%
  \department{School of Computer Science and Engineering}
  \institution{Sun Yat-sen University}
  \city{Guangzhou}
  \country{China}}
\affiliation{%
  \institution{Key Laboratory of Machine Intelligence and Advanced Computing, MOE}
  \city{Guangzhou}
  \country{China}}
\email{wangruix5@mail.sysu.edu.cn}

\begin{abstract}
Out-of-distribution (OOD) detection remains challenging for image classifiers, especially when near-OOD samples lie close to in-distribution (ID) class boundaries. Recent vision-language detectors improve OOD detection through class semantics, local prompting, or LLM-generated outlier concepts, but seldom use language as explicit boundary evidence between confusing ID classes. 
We propose Pairwise Witness Local Rejection (PWLR), which uses an MLLM offline to describe visible local cues that favor one ID class over a specific rival class.
These cue phrases are then screened with ID-only data under a frozen vision-language backbone, so that only reliable local verifiers are kept.
At inference, PWLR first retains a small set of globally plausible classes, then checks whether any of them is locally supported against its most relevant rivals, and finally combines this pairwise local evidence with the global class score through calibration. Experiments on ImageNet-100 far-OOD, cleaner/challenging OOD and near-OOD benchmarks show that PWLR consistently improves strong vision-language baselines across multiple backbones. Source code will be released.
\end{abstract}



\maketitle

\section{Introduction}
\label{sec:intro}

Out-of-distribution (OOD) detection aims to identify test samples that do not belong to the in-distribution (ID) space learned by a model. When deployed in open-world environments, image recognition models often misclassify OOD inputs as ID classes with high confidence, causing potential severe consequences especially for safety-critical fields. 
This problem becomes particularly challenging when near-OOD samples lie close to ID class boundaries, making them difficult to reject using only whole-image class scores.

Most existing OOD detectors either derive post-hoc scores from model confidence or feature geometry~\cite{Hendrycks2016ABF,10.5555/3495724.3497526, pmlr-v162-sun22d}, or improve ID-OOD separation with auxiliary outliers~\cite{hendrycks2018deep}. 
With the rise of pre-trained vision-language models, OOD detection has further moved toward multimodal matching, where images are compared with textual class concepts in a shared embedding space~\cite{NEURIPS2022_e43a3399}. 
Subsequent studies strengthen this paradigm mainly from the text side, for example by introducing negative concepts~\cite{jiang2024neglabel}, expanding the semantic pool of candidate labels~\cite{chen2024csp}, or enriching ID-side textual descriptions beyond plain class names~\cite{lu2025fa}.
Another line complements global matching with local regions, local prompts, or region-aware regularization, showing that local evidence can improve robustness when global representations entangle object and context~\cite{miyai2025zero, miyai2023locoop,zeng2025localprompt}.
More recently, LLM- or MLLM-assisted methods use large models to generate ID descriptions, outlier concepts, or image-adaptive prompts, aiming to inject richer semantic knowledge into the detection process~\cite{dai2023exploring, cao2024eoe, kim2025reguide}.
Despite these advances, language is still used mainly as class semantics, auxiliary concepts, or prompt guidance, rather than as explicit pairwise evidence tied to class competition. This leaves open an important question: can language be converted into stable, visually verifiable boundary evidence that directly supports one plausible class against a confusing rival?

In this paper, we take a different view. Instead of asking an MLLM to imagine what unknown classes may be, we use it offline to describe why one ID class should be preferred over another. 
We propose \textbf{Pairwise Witness Local Rejection (PWLR)}, a boundary-aware OOD detection framework, where the boundary refers to the visual differences that separate one ID class from a confusing rival class. The pipeline uses text to describe these differences and uses local image evidence, rather than only whole-image matching, to verify whether an image supports one class over the rival.
Specifically, for each anchor--rival class pair, we induce a small set of directed witness phrases that describe object-centric local evidence supporting the anchor against the rival.

Crucially, PWLR does not use such witnesses as free-form test-time outputs. Instead, each witness defines a rival-conditioned local hypothesis: if a candidate class is correct against a competing rival, part of the image should exhibit the corresponding anchor-supporting evidence. We test these hypotheses only within the competitive neighborhood of each candidate class, so that local evidence is used to explicitly model class boundaries rather than to provide another generic confidence cue. To make the witness bank quantitatively reliable, we further fit and filter witnesses using ID-only samples under the chosen backbone, converting open-ended language outputs into stable local evidence verifiers.

For the final decision, PWLR calibrates the pairwise local verification score and the global candidate prior to the same scale for each retained candidate class, and then combines them.  Here, the local branch is not used as another generic regional confidence score; instead, it provides directed comparative evidence tied to specific class competition, while the global branch constrains which candidate classes remain semantically plausible at the image level. Consequently, an input is accepted as ID not merely when some class attains a high global match, but when at least one candidate is both globally plausible and locally supported against its most relevant rivals. In this way, PWLR turns OOD detection from direct confidence scoring into calibrated candidate-level acceptance under multimodal comparative boundary evidence.

The main contributions of this work are as follows:
\begin{itemize}
    \item We propose PWLR, a boundary-aware OOD detection framework that reformulates image classification OOD detection as directed pairwise witness verification in the competitive neighbourhood of candidate classes.
    \item We introduce an offline witness induction and ID-only screening pipeline that converts MLLM-generated fine-grained descriptions into stable local evidence verifiers under a frozen VLM backbone.
    \item We further show how PWLR can combine local comparative evidence with a global candidate prior through candidate-level calibration, so that final acceptance requires both global plausibility and local boundary support.
\end{itemize}

\section{Method}

Our framework consists of four sequential stages:
(1) pairwise witness generation,
(2) graph construction and candidate filtering,
(3) pairwise witness local verification, and
(4) retained-class calibrated acceptance.
The first two stages are performed offline to build a reusable witness bank and a competition graph fitted to the chosen backbone, while the last two stages are used online for OOD detection. The ID training data are further divided into two disjoint offline splits: an ID fitting split $\mathcal{D}_{\mathrm{fit}}$ for prototype estimation and ID-only witness screening, and a calibration split $\mathcal{D}_{\mathrm{cal}}$ for retained-class calibration and threshold estimation.
Under the chosen frozen VLM backbone, an image encoder maps an input image $\mathbf{x}$ to a normalized global image embedding $\mathbf{g}(\mathbf{x})$, while a text encoder maps each class label prompt or witness prompt to a normalized text embedding. All similarities in the following sections are computed on normalized embeddings, so inner products are equivalent to cosine similarities. 

\subsection{Pairwise Witness Generation}
\begin{figure}[!t]
  \centering
  \includegraphics[width=\columnwidth]{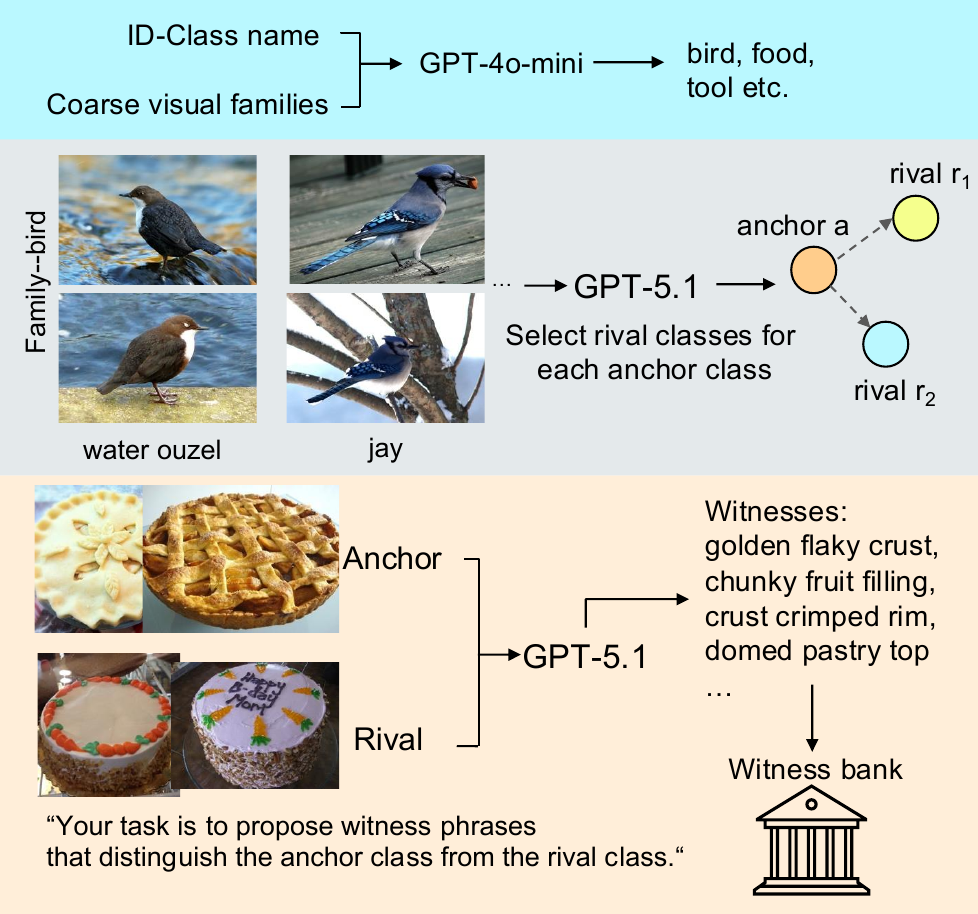}
  \caption{Pipeline for pairwise witness generation. Each ID class is first assigned to a coarse visual family, after which a small set of visually confusable rival classes is selected for each anchor class. For each directed anchor-rival pair, a vision-capable MLLM then generates witness phrases that describe local evidence supporting the anchor against the rival, forming a witness bank for later comparative verification.}
  \Description{A workflow diagram showing how ID class names are first grouped into coarse visual families, how rival classes are selected for each anchor class, and how GPT-5.1 generates directed witness phrases that are stored in a witness bank.}
  \label{fig:witness_generation}
\end{figure}

The first stage (Figure~\ref{fig:witness_generation}) is to construct a pairwise witness bank for later comparative verification using strong multimodal large language models (MLLMs). Rather than querying an MLLM at test time, we use it offline to convert open-ended multimodal knowledge into reusable pairwise evidence tied to class competition. 
Let $\mathcal{C}_{\mathrm{ID}}=\{1,\ldots,M\}$ denote the index set of ID classes, where $M$ is the total number of ID classes, and let $\ell_c$ denote the textual label of class $c\in\mathcal{C}_{\mathrm{ID}}$. For each directed class pair $(a,r)$ with $a,r\in\mathcal{C}_{\mathrm{ID}}$ and $a\neq r$, where $a$ denotes the anchor class and $r$ denotes the rival class, we aim to generate an ordered list of witness phrase candidates for pair $(a,r)$, from which the final witness set is obtained after the later ID-only screening stage, that visually distinguish class $a$ from class $r$, where $L$ is the number of retained witness phrases for pair $(a,r)$ and each $w_i$ is a short natural-language witness phrase. They are not required to share the same length, and they are converted only later by the frozen text encoder into fixed-dimensional text embeddings for scoring.
We use the term witness to emphasize that each phrase functions as directional visual evidence in a pairwise comparison: it supports the anchor class $a$ against a specific rival $r$. Thus, a witness is not a generic caption or attribute, but a class-specific, rivalry-aware cue designed for later comparative verification. Each witness phrase is expected to satisfy two main requirements:
(i) it describes specific visual evidence of the anchor class $a$ that can be seen and examined in the image, and 
(ii) it discriminates against the rival class $r$ which is likely to be visually confusable in natural images. 
For instance, African grey as an anchor class has jacamar as its rival class, considering their similar colorful tail feathers. In the following, we will describe 
how the directed class pairs are instantiated, how rivals are selected for each anchor class, and how the corresponding witness phrases are induced.

\begin{figure*}[!t]
  \centering
  \includegraphics[width=0.9\textwidth]{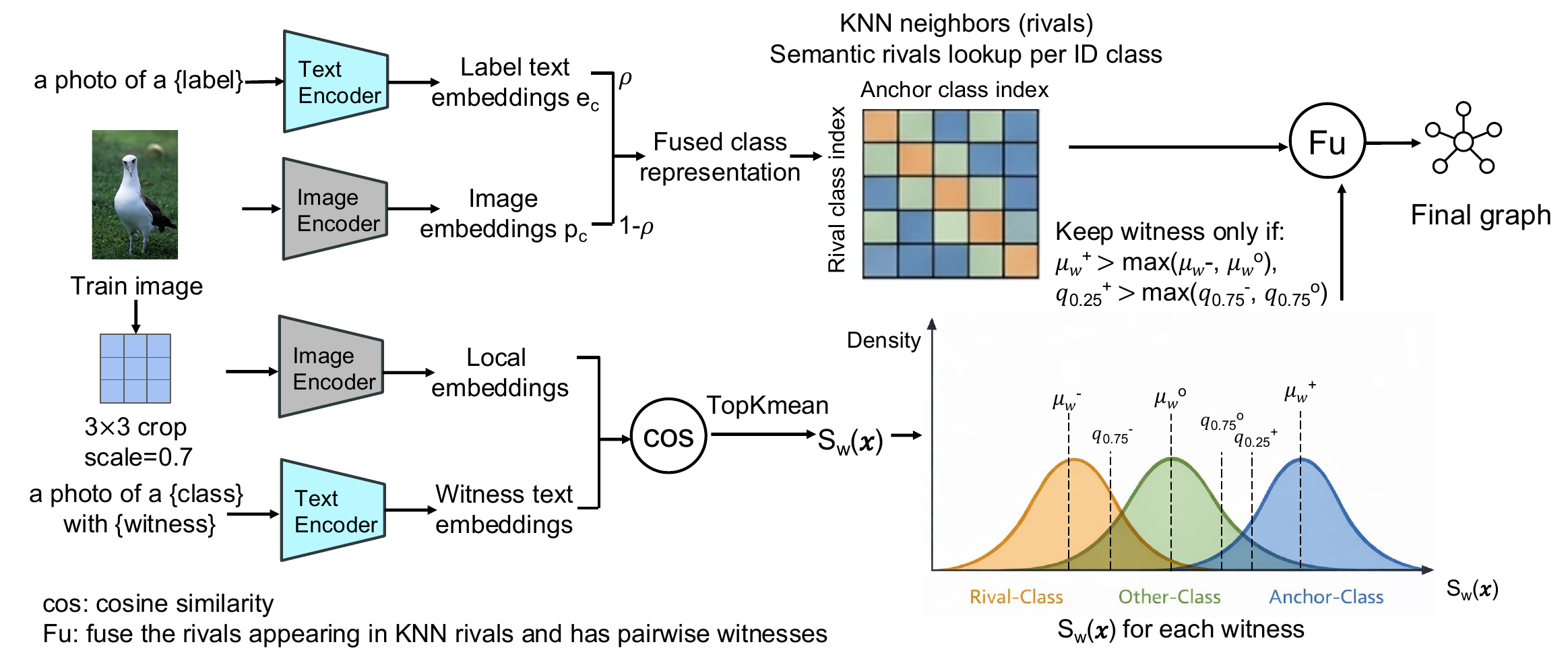}
  \caption{Pipeline for graph construction and filtering. Text representations and class prototypes are fused to define both global class plausibility and semantic rival neighborhoods. Each witness is then fitted on ID samples under the frozen VLM backbone, and only those whose local responses remain consistently higher on anchor-class samples than on rival-class and other-class samples are retained as reliable local verifiers.}
  \label{fig:graph_construction}
\end{figure*}

\noindent \textbf{Coarse visual family assignment.}
Instead of searching rivals from all ID classes directly for any anchor class, each class is first assigned to a coarse visual family. Specifically, for the $c$-th ID class, a GPT model (i.e., GPT-4o-mini) receives the textual label $\ell_c$ and is asked to output a family label, yielding a family mapping $F(c)\in\mathcal{F}$, where $\mathcal{F}$ is a fixed set of 18 coarse visual families, predefined by pruning the WordNet noun hierarchy to mid-level visually coherent object groups, such as bird, dog, building, vehicle, and food, together with a fallback family \texttt{other\_object} for classes not covered by the more specific groups.
This step suppresses implausible cross-family pairings and stabilizes subsequent rival selection when the ID classes candidate catalog is large.

\noindent \textbf{Family-constrained rival selection.}
For an anchor class $a$, the rival search space is restricted to the same-family rival pool $\mathcal{R}(a)=\{c\in\mathcal{C}_{\mathrm{ID}} \mid F(c)=F(a),\ c\neq a\}$.
Since the GPT model does not consume a mathematical set directly, $\mathcal{C}(a)$ is further introduced to denote the candidate catalog obtained by serializing $\mathcal{R}(a)$ into a structured list of class entries. Thus, $\mathcal{C}(a)$ is the prompt-level representation of $\mathcal{R}(a)$ rather than a different search space. The model then receives the anchor label $\ell_a$, the candidate catalog $\mathcal{C}(a)$, and two supporting images randomly sampled from class $a$ in the ID training set, and is prompted as follows: \begin{quote}
\footnotesize\itshape
You are selecting visually confusable rival classes for image classification. Choose exactly five rival classes from candidate pool. Prefer classes that are likely to be confused with the anchor in natural images and have similar visible structure, shape, parts, texture, or coarse appearance. Use the support images to focus on actual visual morphology rather than lexical relatedness. Do not choose the anchor itself. Return only the selected rival classes.
\end{quote}
The resulting rival neighbor set is denoted by $\mathcal{N}(a)\subseteq\mathcal{R}(a)$. In the current implementation, a small fixed number of rivals is retained for each anchor class.

\noindent \textbf{Pairwise witness generation.}
For each directed pair $(a,r)$ with $r\in\mathcal{N}(a)$, supporting images from both classes are provided to GPT-5.1, a vision-capable multimodal large language model, which is then asked to propose short textual phrases for visual description (i.e., witnesses) that support class $a$ against class $r$.
The model is instructed to describe local visual evidence rather than backgrounds or generic captions. To facilitate later quality control, each witness is also accompanied by two discrete attributes: a visibility score and a context-only risk score, both on a 1--3 scale.
The visibility score measures whether the phrase corresponds to a clear and visually identifiable local cue on the foreground object associated with class $a$, such as a distinctive part, shape detail, texture pattern, or local arrangement, that can be recognized from a limited image region, whereas the context-only risk score measures whether the phrase could be recognized largely from surrounding context instead of intrinsic object itself. 
In the current configuration, two anchor images and two rival images are randomly sampled from the ID training set, twelve primary phrases are requested in each round, and three rounds with the same prompt template are run for each directed pair $(a,r)$.

\noindent \textbf{Robust parsing and witness finalization.}
After merging duplicate phrases across rounds, each surviving candidate phrase $p$ is summarized by three statistics: its occurrence count $n(p)$ across the three rounds, its accumulated visibility score $v(p)$, and its accumulated context-risk score $u(p)$. To keep this stage simple and avoid mixing incomparable quantities, we order the candidates lexicographically: we first sort them by $n(p)$ in descending order; ties are then broken by the average visibility $\bar{v}(p)=v(p)/n(p)$ in descending order; any remaining ties are finally broken by the average context-risk $\bar{u}(p)=u(p)/n(p)$ in ascending order. We do not introduce an additional truncation hyperparameter at this stage. Instead, the resulting ordered candidate list is stored in the pairwise witness bank and is later screened with ID-only data in Section~\ref{subsec:graph_construction}. The witness bank is stored in JSONL format; for each directed pair $(a,r)$, it contains the ordered merged candidates with their summary statistics and the raw outputs from the three prompting rounds.
This pairwise witness bank is then used subsequently to construct the directed witness graph (Section~
\ref{subsec:graph_construction}) and for online pairwise local verification (Section~\ref{subsec:verify}), so no MLLM call is required during test-time inference.

\subsection{Graph Construction and Filtering}
\label{subsec:graph_construction}
After the pairwise witness bank is generated, a directed witness graph is constructed to organize the selected rival relations and their associated witness sets for later pairwise local verification (Figure~\ref{fig:graph_construction}). 
The nodes are ID classes, and a directed edge $(a,r)$ indicates that class $r$ is treated as a selected rival of anchor class $a$. Each edge stores the witness set $\mathcal{W}(a\!\to\! r)$. This graph specifies which rival comparisons are available for each ID class and which witness set is attached to each comparison. The later pairwise local verification stage (Section~\ref{subsec:verify}) then evaluates whether an input image contains local visual evidence supporting one class against a specific rival, rather than assigning a single undirected compatibility score to the whole image.

\noindent \textbf{Text embeddings and global class scoring.}
For each class $c$, a small prompt ensemble is used to obtain a stable text representation of the class semantics, rather than relying on a single wording that may bias the text embedding. Specifically, we instantiate three class templates with the textual label $\ell_c$:
``a photo of a \{label\}'',
``a close-up photo of a \{label\}'',
and ``a centered photo of a \{label\}'',
and the resulting text embeddings are averaged to obtain the class representation $\mathbf{e}_c$.
Only three templates are used because the goal is not to build a large prompt ensemble, but to reduce sensitivity to any single phrasing while still keeping the text representation simple and stable. These three templates cover complementary object views: a generic class description, a close-up view that better matches fine-grained local detail, and a centered object-centric view that suppresses peripheral context.
For each witness phrase $w$ on the edge $(a,r)$, three witness templates are similarly instantiated with the anchor class label $\ell_a$, where \{anchor\} denotes the textual label of the anchor class $a$:
``a photo of a \{anchor\} with \{phrase\}'',
``a \{anchor\} showing \{phrase\}'',
and ``\{anchor\}, \{phrase\}'',
The resulting embeddings are averaged to obtain the witness representation $\mathbf{t}_{a,r,w}$.

The same design principle is used here: multiple phrasings make the witness representation less brittle to prompt wording while keeping it anchored to the directed comparison $a\rightarrow r$.
To score whole-image plausibility and define nearby rival classes before local verification, class prototypes are further introduced.
Let $\mathbf{p}_c$ denote the class prototype of class $c$, defined as the mean of normalized global image embeddings over class-$c$ images in the ID fitting split $\mathcal{D}_{\mathrm{fit}}$.
The same ID fitting split is used for ID-only witness screening.
For a directed pair $(a,r)$, witness screening uses anchor-class samples from class $a$, rival-class samples from class $r$, and other-class samples from all remaining ID classes $\mathcal{C}_{\mathrm{ID}}\setminus\{a,r\}$.
Using only the class text representation $\mathbf{e}_c$ would preserve class semantics but ignore the empirical image distribution of class $c$, whereas using only the class prototype $\mathbf{p}_c$ would capture the image distribution but discard the text-side semantic prior. The two are therefore combined to define the global class score of class $c$ for input image $\mathbf{x}$:
\begin{equation}
G_c(\mathbf{x})=(1-\rho)\,\mathbf{g}(\mathbf{x})^\top \mathbf{e}_c+\rho\,\mathbf{g}(\mathbf{x})^\top \mathbf{p}_c \,,
\end{equation}
where $\mathbf{g}(\mathbf{x})$ is the normalized global image embedding of $\mathbf{x}$ under the frozen vision-language backbone, and $\rho$ balances the text and prototype terms. The score $G_c(\mathbf{x})$ measures the global plausibility of class $c$ for image $\mathbf{x}$ and is used for retained-class selection.
The proposed pipeline retains the top-$K_g$ (e.g., 5) ID classes with the largest $G_c(\mathbf{x})$ values and denotes them by the retained class set $\mathcal{Y}(\mathbf{x})$. Subsequent pairwise witness verification is performed only for classes in $\mathcal{Y}(\mathbf{x})$. 
This truncation avoids spending pairwise verification budget on clearly unreasonable ID classes and focuses the subsequent witness-based comparisons on a small set of globally plausible ID classes.

\noindent \textbf{Semantic-pruned rival subgraph.}
The raw witness graph is constructed to provide sufficient rival coverage for each ID class, but not every stored rival edge is equally relevant at test time. Therefore, the pipeline does not compare each retained ID class $y \in \mathcal{Y}(\mathbf{x})$ against all of its outgoing rival classes in the raw witness graph. 
Comparing against all rivals would introduce unnecessary computation and may include semantically distant or weakly relevant rival classes in the subsequent pairwise local witness verification stage.
Thus, a semantic-pruned rival subgraph is built for each ID class.
Specifically, a fused class representation $\mathbf{h}_c=\text{Norm} ((1-\rho) \mathbf{e}_c + \rho \mathbf{p}_c )$ is formed, where the same coefficient $\rho$ is used to combine text and prototype representations, so that global class scoring and rival pruning are defined in compatible spaces. Pairwise class similarities are then computed by $\mathbf{h}_c^\top \mathbf{h}_{c'}$, and a small semantic-neighbor set (e.g., 12) is retained for each ID class. This semantic-neighbor set preserves plausible rival classes while excluding clearly irrelevant ones.

For each retained ID class $y \in \mathcal{Y}(\mathbf{x})$, up to two outgoing rival classes are retained. A rival class $j$ is kept only if the directed edge $(y,j)$ exists in the witness graph and class $j$ belongs to the retained semantic-neighbor set of $y$. This second pruning step restricts pairwise witness verification to the most relevant nearby rival classes of $y$, rather than all rival classes linked offline. If no outgoing rival class of $y$ satisfies this condition, the single most similar outgoing rival class is retained as fallback. The resulting active rival set is denoted by $\mathcal{J}(y)$.



\noindent \textbf{ID-only witness screening.}
The goal of ID-only witness screening is to retain only those witnesses whose image-level responses are consistently higher on anchor-class samples than on rival-class and other-class samples under the frozen vision-language backbone. After the active rival set is determined, each witness on an active edge is evaluated on $\mathcal{D}_{\mathrm{fit}}$.
This step checks whether an MLLM-generated witness phrase yields stronger image-level responses on anchor-class samples than on non-anchor samples under the frozen backbone.

Each image is represented by $T=9$ local embeddings extracted from a $3\times3$ crop grid. An explicit crop-grid representation is used instead of model-specific internal patch tokens so that witness screening remains comparable across different vision-language backbones while still providing a small but diverse set of local views.
Given the witness representation $\mathbf{t}_{a,r,w}$ and the local embeddings $\{\mathbf{l}_t(\mathbf{x})\}_{t=1}^{T}$, local-view similarities are computed by $s_{t,w}(\mathbf{x})=\operatorname{cos}(\mathbf{l}_t(\mathbf{x}),\mathbf{t}_{a,r,w})$, and the image-level witness response is defined as
\begin{equation}
s_w(\mathbf{x})=\operatorname{TopKMean}\!\left(\{s_{t,w}(\mathbf{x})\}_{t=1}^{T},k_p\right) \,,
\end{equation}
where $\operatorname{TopKMean}$ averages the largest $k_p$ local-view similarities. This design reflects the assumption that a valid witness usually appears only in part of the object, so using only the strongest few local views preserves localized evidence while reducing dilution from irrelevant regions.

For each witness $w$ on edge $(a,r)$, the response $s_w(\mathbf{x})$ is evaluated on anchor-class, rival-class, and other-class screening samples, where the other-class fitting samples come from the remaining ID classes $\mathcal{C}_{\mathrm{ID}}\setminus\{a,r\}$. From these three response sets, the means $\mu_w^{+}$, $\mu_w^{-}$, and $\mu_w^{o}$, the pooled standard deviation $\sigma_w$, and the quantiles $q_{0.25}^{+}$, $q_{0.75}^{-}$, and $q_{0.75}^{o}$ are estimated. 
Let $\mu_w^{\mathrm{ref}}=\max(\mu_w^{-},\mu_w^{o})$. Then the effect size of witness $w$ is defined as $e_w=(\mu_w^{+}-\mu_w^{\mathrm{ref}})/\sigma_w$, which measures average anchor-versus-non-anchor separation, while $g_w=q_{0.25}^{+}-\max(q_{0.75}^{-},q_{0.75}^{o})$ measures whether this separation remains under harder tail cases. A witness is kept only if $\mu_w^{+}>\mu_w^{\mathrm{ref}}$ and $q_{0.25}^{+}>\max(q_{0.75}^{-},q_{0.75}^{o})$.
For each kept witness, the midpoint $m_w=\frac{1}{2}(\mu_w^{+}+\mu_w^{\mathrm{ref}})$ is stored for later standardization. 
The standardized score is centered at $m_w$, which lies between the anchor-class mean and the stronger non-anchor mean. It therefore reflects comparative support rather than absolute similarity.

\begin{figure}[!t]
  \centering
  \includegraphics[width=\columnwidth]{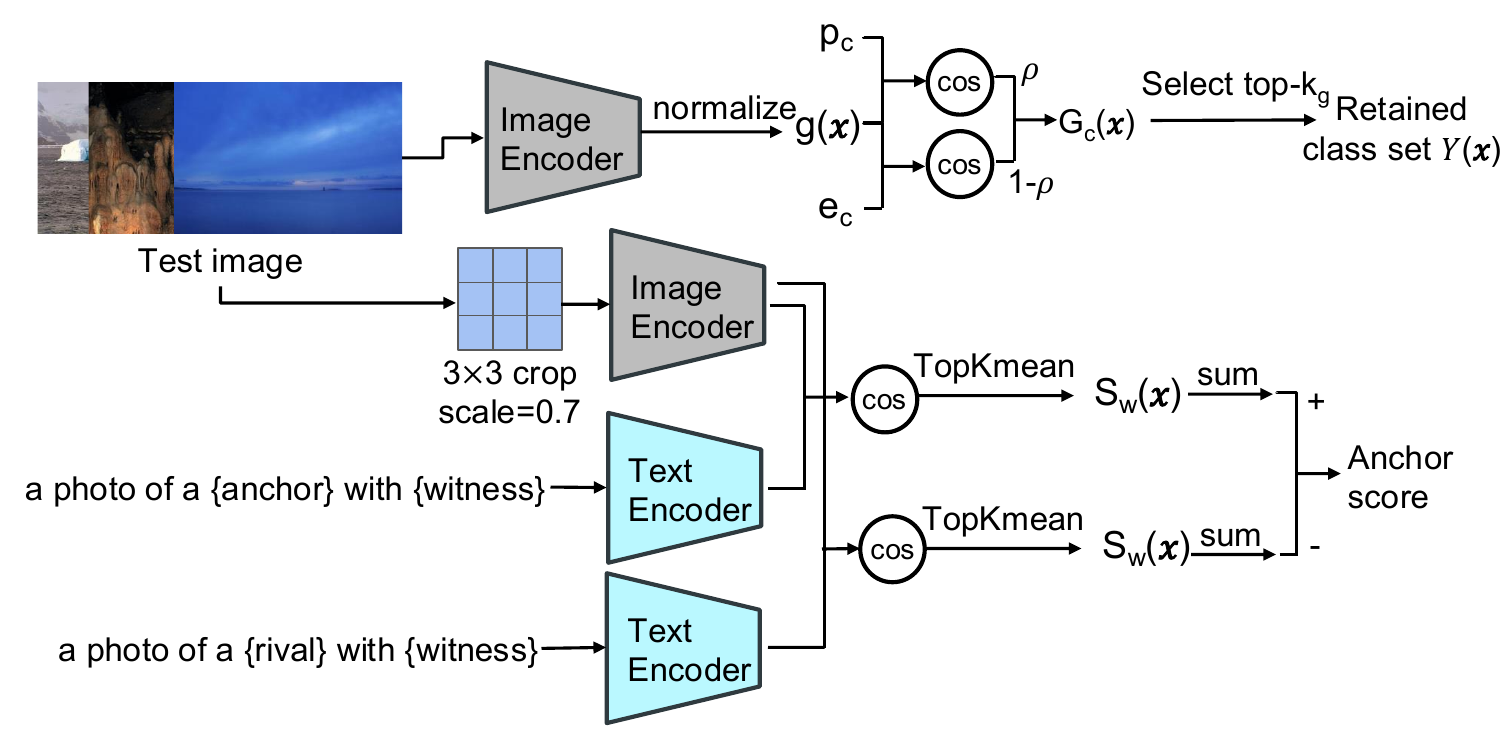}
  \caption{Pipeline for pairwise local witness verification. Given a test image, the pipeline first retains a small set of globally plausible classes using the fused global score, and then verifies each retained class only against its active rivals in the semantic-pruned witness graph. For each directed class pair, witness responses are computed from local views, standardized, and aggregated into a directed support score.}
  \label{fig:pairwise_verification}
\end{figure}

\subsection{Pairwise Local Witness Verification}
\label{subsec:verify}

Given an input image $\mathbf{x}$, a retained candidate class $y\in\mathcal{Y}(\mathbf{x})$, and an active rival class $j\in\mathcal{J}(y)$, the retained witness set $\mathcal{W}(y\!\to\!j)$ is used to evaluate whether $\mathbf{x}$ contains the local visual evidence that supports $y$ against $j$. This stage forms the core local verification step (Figure~\ref{fig:pairwise_verification}). Rather than assessing whether $\mathbf{x}$ is generally compatible with class $y$, the verification focuses on whether the image exhibits the specific local evidence that distinguishes $y$ from the particular rival $j$. Here, the boundary is instantiated as the directed distinction between $y$ and $j$, rather than as an abstract global separation score.

For each retained witness $w\in\mathcal{W}(y\!\to\!j)$, the image-level witness response $s_w(\mathbf{x})$ defined in the screening stage is first computed and then standardized by
\begin{equation}
z_w(\mathbf{x})=\frac{s_w(\mathbf{x})-m_w}{\sigma_w+\varepsilon} \,,
\end{equation}
where $m_w=\frac{1}{2}(\mu_w^{+}+\mu_w^{\mathrm{ref}})$ is the midpoint between the anchor-class mean and the stronger non-anchor mean for witness $w$, $\sigma_w$ is the pooled standard deviation estimated from the anchor-class, rival-class, and other-class screening samples, and $\varepsilon$ is a small constant for numerical stability. This standardization is necessary because different witnesses may have very different raw similarity scales under the frozen encoder. Conversion to a common centered scale makes responses from different witnesses more comparable.
\begin{equation}
\phi_{y,j}(\mathbf{x})=
\frac{\sum_{w\in\mathcal{W}(y\!\to\!j)} \alpha_w z_w(\mathbf{x})}
{\sum_{w\in\mathcal{W}(y\!\to\!j)} \alpha_w} \,,
\qquad
\alpha_w=\max(e_w,0) \,,
\end{equation}
where $e_w$ is the fitted effect size of witness $w$. Because only witnesses that pass the ID-only screening criteria are retained, the corresponding weights satisfy $\alpha_w>0$, so the normalization is well-defined. Larger weights are therefore assigned to witnesses that showed stronger anchor-versus-reference separation during screening.

If the reverse edge $(j,y)$ also exists, it is used only as a one-sided penalty:
\begin{equation}
r_{y,j}(\mathbf{x})=\phi_{y,j}(\mathbf{x})-[\phi_{j,y}(\mathbf{x})]_{+} \,,
\end{equation}
where $[\cdot]_{+}=\max(\cdot,0)$. If the reverse edge is absent, $r_{y,j}(\mathbf{x})=\phi_{y,j}(\mathbf{x})$ is used directly. Positive evidence on $(y,j)$ is treated as direct support for class $y$, whereas positive evidence on $(j,y)$ only reduces the support for $y$ and is not treated as positive support for $y$ itself.
For each retained class $y\in\mathcal{Y}(\mathbf{x})$, the pairwise local verification score is defined as $A_y(\mathbf{x})=\min_{j\in\mathcal{J}(y)} r_{y,j}(\mathbf{x})$, which is the smallest penalized directed support for class $y$ over its active rival set. Therefore, a high value of $A_y(\mathbf{x})$ is obtained only when support for class $y$ remains strong against every active rival in $\mathcal{J}(y)$.
Finally, the image-level pairwise verification score is $S_{\mathrm{ver}}(\mathbf{x})=\max_{y\in\mathcal{Y}(\mathbf{x})}A_y(\mathbf{x})$, and the retained class with the maximum score is taken as the predicted ID label of the local verification branch.

\subsection{Retained-Class Calibrated Acceptance}
\label{subsec:final_acceptance}

Although $A_y(\mathbf{x})$ captures pairwise local support for a retained class $y$, local verification alone is not sufficient for final acceptance. Strong local support may still arise from a small set of local cues even when class $y$ is globally implausible under the global class score $G_y(\mathbf{x})$. A final retained-class calibration is therefore applied to combine local support and global plausibility on a common scale.

Using the calibration split $\mathcal{D}_{\mathrm{cal}}$, one score pair is collected for every retained class of every image in $\mathcal{D}_{\mathrm{cal}}$. Each pair is $(G_y(\mathbf{x}),A_y(\mathbf{x}))$, where $\mathbf{x}\in\mathcal{D}_{\mathrm{cal}}$ and $y\in\mathcal{Y}(\mathbf{x})$. Here, $G_y(\mathbf{x})$ is the global class score from retained-class selection, and $A_y(\mathbf{x})$ is the pairwise local verification score defined above. 
The empirical cumulative distribution functions of the scalar sets $\{G_y(\mathbf{x})\mid \mathbf{x} \in\mathcal{D}_{\mathrm{cal}},\, y\in\mathcal{Y}(\mathbf{x})\}$ and $\{A_y(\mathbf{x})\mid \mathbf{x} \in\mathcal{D}_{\mathrm{cal}},\, y\in\mathcal{Y}(\mathbf{x})\}$ are then computed, denoted by $F_G(\cdot)$ and $F_A(\cdot)$, respectively. This maps the two scores to a common rank-based scale, because their raw magnitudes are not directly comparable.

\begin{table*}[!t]
\centering
\small
\setlength{\tabcolsep}{4.0pt}
\caption{OOD detection performance (\%) on the conventional far-OOD benchmark, with ImageNet-100 as ID dataset and 
iNaturalist, SUN, Places, Textures as OOD datasets. 
The best and second-best results are shown in bold and underline, respectively.}
\label{tab:main_far_ood}
{%
\begin{tabular}{lcccccccccc}
\toprule
\multirow{2}{*}{Method} & \multicolumn{2}{c}{iNaturalist} & \multicolumn{2}{c}{SUN} & \multicolumn{2}{c}{Places} & \multicolumn{2}{c}{Texture} & \multicolumn{2}{c}{Average} \\
\cmidrule(lr){2-3} \cmidrule(lr){4-5} \cmidrule(lr){6-7} \cmidrule(lr){8-9} \cmidrule(lr){10-11}
 & FPR95$\downarrow$ & AUROC$\uparrow$ & FPR95$\downarrow$ & AUROC$\uparrow$ & FPR95$\downarrow$ & AUROC$\uparrow$ & FPR95$\downarrow$ & AUROC$\uparrow$ & FPR95$\downarrow$ & AUROC$\uparrow$ \\
\midrule
MCM & 3.72 & 98.86 & 13.94 & 96.00 & 18.52 & 94.85 & 19.06 & 95.24 & 13.81 & 96.24 \\
GL-MCM & 3.00 & 99.22 & 13.50 & 96.53 & 17.62 & 95.46 & 17.20 & 95.93 & 12.83 & 96.78 \\
MSP & 13.74 & 95.99 & 20.10 & 94.23 & 25.62 & 93.53 & 11.98 & 97.43 & 17.86 & 95.30 \\
Energy & 7.32 & 97.90 & 16.44 & 95.33 & 22.50 & 94.77 & 8.64 & 97.70 & 13.73 & 96.43 \\
ReAct & 6.04 & 98.29 & 13.94 & 96.00 & 20.86 & 95.14 & 8.82 & 97.58 & 12.41 & 96.75 \\
MaxLogit & 8.20 & 97.53 & 16.92 & 95.30 & 22.72 & 94.68 & 8.74 & 97.86 & 14.14 & 96.34 \\
NegLabel & \textbf{0.02} & \textbf{99.97} & 10.04 & 97.83 & 16.10 & 96.02 & 10.30 & 97.44 & 9.12 & 97.82 \\
DPM-F & 6.52 & 98.20 & 21.04 & 93.64 & 27.06 & 92.80 & 15.02 & 96.58 & 17.41 & 95.30 \\
AdaNeg & 0.16 & 99.94 & 9.70 & 97.73 & 19.58 & 96.15 & 10.84 & 97.50 & 10.07 & 97.83 \\
Mahalanobis & 15.38 & 93.35 & 25.00 & 89.00 & 31.00 & 87.32 & 8.82 & 97.62 & 20.05 & 91.82 \\
KNN & 20.70 & 89.45 & 27.16 & 88.62 & 30.36 & 88.61 & 16.02 & 94.98 & 23.56 & 90.42 \\
NNGuide & 1.40 & 99.57 & 11.82 & 97.05 & 19.34 & 95.98 & \underline{8.22} & 98.08 & 10.20 & 97.67 \\
CSP & \underline{0.04} & 99.94 & 8.20 & \underline{98.14} & 14.74 & 96.72 & 8.84 & \underline{98.13} & 7.96 & \underline{98.23} \\
SeTAR & 9.94 & 97.75 & 20.72 & 93.83 & 26.50 & 92.53 & 19.72 & 94.65 & 19.22 & 94.69 \\
\midrule
PWLR-NegLabel & 0.10 & 99.94 & \underline{7.50} & 98.05 & \underline{11.12} & \underline{97.06} & 8.68 & 97.85 & \underline{6.85} & 98.23 \\
PWLR-CSP& \underline{0.04} & \underline{99.96} & \textbf{6.86} & \textbf{98.36} & \textbf{9.74} & \textbf{97.55} & \textbf{7.16} & \textbf{98.31} & \textbf{5.95} & \textbf{98.55} \\
\bottomrule
\end{tabular}%
}
\end{table*}

For each retained class $y\in\mathcal{Y}(\mathbf{x})$, the calibrated global and local scores are defined as $u_G(\mathbf{x},y)=F_G(G_y(\mathbf{x}))$ and $u_A(\mathbf{x},y)=F_A(A_y(\mathbf{x}))$, respectively. The calibrated acceptance score of the retained class $y$ is then defined as 
\begin{equation}
C_y(\mathbf{x})=\lambda_a\,u_A(\mathbf{x},y)+(1-\lambda_a)\,u_G(\mathbf{x},y) \,,
\end{equation}
where $\lambda_a$ balances calibrated local support and calibrated global plausibility.
The final image-level acceptance score is $S_{\mathrm{PWLR}}(\mathbf{x})=\max_{y\in\mathcal{Y}(\mathbf{x})} C_y(\mathbf{x})$, and the maximizing retained class is taken as the predicted ID label. Thus, the final decision is determined by the strongest retained-class hypothesis after joint calibration of the local and global branches.
$S_{\mathrm{PWLR}}(\mathbf{x})$ also serves as the final image-level ID score for OOD detection, with larger values indicating that $\mathbf{x}$ is more likely in-distribution. 

\section{Experiment}
\subsection{Experimental Setup}
\noindent \textbf{Datasets.}
Following previous studies on vision-language OOD detection~\cite{Huang2021MOSTS}, we adopt the ImageNet-100 benchmark~\cite{5206848,NEURIPS2022_e43a3399}. We evaluate on four standard far-OOD datasets, including iNaturalist~\cite{Horn2017TheIS}, SUN~\cite{5539970}, Places~\cite{7968387}, and Textures~\cite{6909856}. To further assess robustness under cleaner and more realistic distribution shifts, we additionally report results on OpenImage-O~\cite{9879414}, ImageNet-O~\cite{hendrycks2021nae}, and NINCO~\cite{pmlr-v202-bitterwolf23a}, following recent challenging OOD evaluation protocols. Beyond ImageNet-100, we further validate our method on several additional fine-grained 
datasets, including CUB-200-2011~\cite{WahCUB_200_2011}, Oxford-IIIT Pet~\cite{6248092}, and Food-101~\cite{10.1007/978-3-319-10599-4_29}. In particular, following~\cite{cao2024eoe}, CUB-200-2011 is partitioned into two disjoint subsets, with one CUB-100 split used as ID classes and the other CUB-100 split used as OOD classes; Oxford-IIIT Pet is similarly partitioned into 18 ID classes and 19 OOD classes. These class-partitioned settings simulate more challenging semantic shifts, because the OOD classes remain visually and semantically close to the ID classes.

\noindent \textbf{Implementation details.}
Unless otherwise specified, all main results are obtained with SigLIP2 ViT-L/16-256 (WebLI)~\cite{big_vision, zhai2023sigmoid, Tschannen2025SigLIP2M} as the default frozen backbone. Cross-backbone generality is further evaluated with PE-Core-L/14-336~\cite{bolya2025PerceptionEncoder, cho2025PerceptionLM} and DFN2B-CLIP ViT-L/14~\cite{fang2023data}. For PWLR, each ID dataset is further divided into an offline ID fitting split and a calibration split. Witness induction is performed offline once for each ID dataset; prototype estimation and witness screening use the ID fitting split, while retained-class calibration uses the calibration split.
The default hyperparameter settings in the main experiments are $\lambda_a=0.4$, $\rho=0.35$, $K_g=5$ and $k_p=2$. 
For fair comparison, all methods use the same dataset splits, image preprocessing pipeline, evaluation protocol, and backbone under each comparison setting.

\noindent \textbf{Comparative Methods.}
We compare our method against a diverse set of representative OOD detectors. These include generic post-hoc baselines, including MSP~\cite{Hendrycks2016ABF}, Energy~\cite{10.5555/3495724.3497526}, ReAct~\cite{NEURIPS2021_01894d6f}, MaxLogit~\cite{pmlr-v162-hendrycks22a}, Mahalanobis~\cite{10.5555/3327757.3327819}, KNN~\cite{pmlr-v162-sun22d}, and NNGuide~\cite{park2023nearest}; vision-language concept-matching methods such as MCM~\cite{NEURIPS2022_e43a3399} and GL-MCM~\cite{miyai2025zero}; and recent semantics-enhanced or VLM-based baselines, including NegLabel~\cite{jiang2024neglabel}, CSP~\cite{chen2024csp}, DPM-F~\cite{10.1007/978-3-031-73013-9_16}, AdaNeg~\cite{NEURIPS2024_4462db5e}, and SeTAR~\cite{NEURIPS2024_8555cf30}. This selection covers both classical confidence-based detectors and stronger recent VLM-based approaches, enabling a comprehensive evaluation under a unified protocol. For all compared methods, we use the same backbone and the same ID-OOD splits as our method.

\noindent \textbf{Metrics.}
We follow standard OOD detection protocols and use FPR95 and AUROC as the primary metrics. Specifically, FPR95 denotes the false positive rate on OOD samples when the true positive rate on ID samples is fixed at 95\%, and thus lower is better. AUROC measures threshold-free separability between ID and OOD scores over the entire operating range, and thus higher is better. For each benchmark, we report results on individual OOD datasets and their average. Following standard OOD evaluation on fixed dataset splits, the main tables report one run per benchmark setting. Since the proposed pipeline uses offline witness generation before evaluation, the reported results correspond to one instantiated witness bank under each setting; the stability under repeated witness generation is further discussed in the appendix.

\subsection{Main Results}

\noindent \textbf{Results on the conventional far-OOD benchmark.}
Table~\ref{tab:main_far_ood} reports the conventional far-OOD results. PWLR-CSP achieves the best average performance at 5.95\% FPR95 and 98.55\% AUROC, while PWLR-NegLabel achieves 6.85\% FPR95 and 98.23\% AUROC. The gains are particularly clear on Places and remain consistent on SUN and Textures, whereas the margin on iNaturalist is smaller because several strong baselines are close to saturation. These results show that pairwise local witness verification is most helpful when global matching leaves substantial ambiguity among competitive ID classes.

\begin{table}[!t]
\centering
\small
\setlength{\tabcolsep}{4.0pt}
\caption{OOD detection performance  on the 
more challenging OOD benchmark,  with ImageNet-100 as ID dataset and 
OpenImage-O, ImageNet-O, NINCO as OOD datasets. 
}
\label{tab:main_clean_ood}
\resizebox{\columnwidth}{!}{%
\begin{tabular}{lcccccccc}
\toprule
\multirow{2}{*}{Method} & \multicolumn{2}{c}{OpenImage-O} & \multicolumn{2}{c}{ImageNet-O} & \multicolumn{2}{c}{NINCO} & \multicolumn{2}{c}{Average} \\
\cmidrule(lr){2-3} \cmidrule(lr){4-5} \cmidrule(lr){6-7} \cmidrule(lr){8-9}
 & FPR95$\downarrow$ & AUROC$\uparrow$ & FPR95$\downarrow$ & AUROC$\uparrow$ & FPR95$\downarrow$ & AUROC$\uparrow$ & FPR95$\downarrow$ & AUROC$\uparrow$ \\
\midrule
MCM & 9.10 & 97.69 & 16.16 & 95.83 & 29.64 & 94.90 & 18.30 & 96.14 \\
GL-MCM & 7.96 & 98.17 & 14.76 & 96.33 & 25.52 & 95.44 & 16.08& 96.65\\
MSP & 13.24 & 96.84 & 18.36 & 95.62 & 24.18 & 94.69 & 18.59 & 95.72 \\
Energy & 10.38 & 97.09 & 15.38 & 96.18 & 26.78 & 94.23 & 17.51 & 95.83 \\
ReAct & 8.28 & 97.45 & 13.42& 96.52 & 28.16 & 94.17 & 16.62 & 96.05 \\
MaxLogit & 10.46 & 97.30 & 15.60 & 96.27 & 26.64 & 94.58 & 17.57 & 96.05 \\
NegLabel & 10.64 & 97.93 & 16.46 & 96.36 & 20.98 & 95.31 & 16.03 & 96.53 \\
DPM-F & 12.84 & 97.01 & 15.06 & 96.72& 24.54 & 93.93 & 17.48 & 95.89 \\
AdaNeg & 11.16 & 97.69 & 24.54 & 94.79 & 21.38 & 94.94 & 19.03 & 95.81 \\
Mahalanobis & 7.54 & 97.38 & 17.28 & 95.20 & 28.40 & 90.87 & 17.74 & 94.48 \\
KNN & 12.10 & 96.32 & 25.16 & 93.04 & 49.30 & 84.17 & 28.85 & 91.18 \\
NNGuide & 11.52 & 97.10 & 17.32 & 96.03 & \underline{18.48} & \underline{96.06} & 15.77& 96.40 \\
CSP & 12.20 & 97.41 & 17.42 & 95.98 & 23.74 & 94.19 & 17.79 & 95.86 \\
SeTAR & 14.52 & 96.38 & 22.98 & 94.49 & 25.56 & 93.41 & 21.02 & 94.76 \\
\midrule
PWLR-NegLabel& \textbf{6.72} & \textbf{98.60} & \textbf{12.56} & \underline{97.16}& \textbf{18.30} & \textbf{96.08} & \textbf{12.53} & \textbf{97.28} \\
PWLR-CSP& \underline{7.28} & \underline{98.47} & \underline{12.66}& \textbf{97.17}& 20.18 & 95.70 & \underline{13.37}& \underline{97.11} \\
\bottomrule
\end{tabular}%
}
\end{table}

\noindent \textbf{Results on the cleaner/challenging OOD benchmark.}
Table~\ref{tab:main_clean_ood} reports the results on the more challenging benchmark, where the gains of PWLR become larger. PWLR-CSP improves CSP from 17.79\%/95.86\% to 13.37\%/97.11\% in average FPR95/AUROC, and PWLR-NegLabel improves NegLabel from 16.03\%/96.53\% to 12.53\%/ 97.28\%. The improvement is consistent across OpenImage-O, ImageNet-O, and NINCO, supporting that PWLR improves class-boundary discrimination rather than relying on coarse dataset-specific cues.

\begin{table}[t]
\centering
\scriptsize
\setlength{\tabcolsep}{2.8pt}
\renewcommand{\arraystretch}{1.05}
\caption{Multi-backbone results of different plug-in variants on the split near-OOD benchmarks. We report FPR95/AUROC on Pet18/19 and CUB100/100 under three frozen vision-language backbones. 
}
\label{tab:multi_backbone_split_plugins}
\resizebox{\columnwidth}{!}{%
\begin{tabular}{llcccc}
\toprule
\multirow{2}{*}{Architecture} & \multirow{2}{*}{Method} & \multicolumn{2}{c}{Pet18/19} & \multicolumn{2}{c}{CUB100/100} \\
\cmidrule(lr){3-4}\cmidrule(lr){5-6}
& & FPR95$\downarrow$ & AUROC$\uparrow$ & FPR95$\downarrow$ & AUROC$\uparrow$ \\
\midrule

\multirow{8}{*}{SigLIP2-L16-256}
& MCM & 38.01 & 93.48 & 63.72 & 84.85 \\
& \textbf{PWLR-MCM} & \textbf{27.77} & \textbf{94.99} & \textbf{60.07} & \textbf{85.29} \\
\cmidrule(lr){2-6}
& MSP & \textbf{27.02} & 94.07 & 73.14 & 78.24 \\
& \textbf{PWLR-MSP} & 29.91 & \textbf{94.81} & \textbf{70.61} & \textbf{80.76} \\
\cmidrule(lr){2-6}
& NegLabel & 27.19 & 92.81 & 63.30 & 84.40 \\
& \textbf{PWLR-NegLabel} & \textbf{24.96} & \textbf{94.06} & \textbf{61.97} & \textbf{85.09} \\
\cmidrule(lr){2-6}
& CSP & 30.33 & 92.35 & 64.21 & 84.34 \\
& \textbf{PWLR-CSP} & \textbf{24.73} & \textbf{94.23} & \textbf{61.89} & \textbf{85.19} \\
\midrule

\multirow{8}{*}{DFN2B-L14}
& MCM & 33.62 & 92.64 & 43.87 & 91.02 \\
& \textbf{PWLR-MCM} & \textbf{33.60} & \textbf{93.20} & \textbf{43.83} & \textbf{91.30} \\
\cmidrule(lr){2-6}
& MSP & 34.61 & 91.86 & 57.98 & 85.67 \\
& \textbf{PWLR-MSP} & \textbf{33.01} & \textbf{93.37} & \textbf{57.92} & \textbf{87.34} \\
\cmidrule(lr){2-6}
& NegLabel & 43.64 & 91.63 & 48.67 & 90.11 \\
& \textbf{PWLR-NegLabel} & \textbf{38.61} & \textbf{92.00} & \textbf{46.90} & \textbf{90.63} \\
\cmidrule(lr){2-6}
& CSP & 43.05 & 91.53 & 48.04 & 90.34 \\
& \textbf{PWLR-CSP} & \textbf{41.33} & \textbf{92.01} & \textbf{45.80} & \textbf{90.79} \\
\midrule

\multirow{8}{*}{PE-Core-L14-336}
& MCM & 17.47 & 96.57 & 45.70 & 91.01 \\
& \textbf{PWLR-MCM} & \textbf{15.92} & \textbf{96.98} & \textbf{44.39} & \textbf{91.15} \\
\cmidrule(lr){2-6}
& MSP & \textbf{14.54} & 96.70 & 69.40 & 85.21 \\
& \textbf{PWLR-MSP} & 14.60 & \textbf{97.18} & \textbf{59.90} & \textbf{87.47} \\
\cmidrule(lr){2-6}
& NegLabel & 17.16 & 95.89 & 48.70 & 89.13 \\
& \textbf{PWLR-NegLabel} & \textbf{14.18} & \textbf{97.24} & \textbf{40.52} & \textbf{89.31} \\
\cmidrule(lr){2-6}
& CSP & 16.67 & 95.98 & 48.14 & 88.71 \\
& \textbf{PWLR-CSP} & \textbf{13.59} & \textbf{97.34} & \textbf{46.60} & \textbf{89.52} \\
\bottomrule
\end{tabular}%
}
\end{table}

\begin{table}[t]
\centering
\scriptsize
\setlength{\tabcolsep}{2.8pt}
\renewcommand{\arraystretch}{1.05}
\caption{Ablation on the cleaner/challenging benchmarks. (a) Effect of the global prior branch and the pairwise witness verification branch. (b) Effect of two local comparison mechanisms: reverse-edge penalty and active rival pruning. In (b), both core branches are enabled by default.}
\label{tab:ablation_clean_challenging}
\resizebox{\columnwidth}{!}{%
\begin{tabular}{lccc|lccc}
\toprule
\multicolumn{4}{c|}{\textbf{(a) Core branches}} & \multicolumn{4}{c}{\textbf{(b) Local comparison mechanisms}} \\
\cmidrule(lr){1-4}\cmidrule(lr){5-8}
 & \textbf{Global} & \textbf{Pairwise} & \textbf{Full} 
 &  & \textbf{Reverse} & \textbf{Pruning} & \textbf{Full} \\
\midrule
Global Prior          & \checkmark & --         & \checkmark 
& Reverse Penalty     & \checkmark & --         & \checkmark \\
Pairwise Verification & --         & \checkmark & \checkmark 
& Rival Pruning       & --         & \checkmark & \checkmark \\
FPR95 $\downarrow$    & 18.67      & 31.90      & 16.42
& FPR95 $\downarrow$  & 16.64      & 16.90      & 16.42 \\
AUROC $\uparrow$      & 95.13      & 93.69      & 96.23
& AUROC $\uparrow$    & 96.23      & 96.20      & 96.44 \\
\bottomrule
\end{tabular}%
}
\end{table}

\noindent \textbf{Results on the split near-OOD benchmarks and multi-backbone validation.}
Table~\ref{tab:multi_backbone_split_plugins} reports the split near-OOD results under three vision-language backbones. Across 24 plug-in comparisons, PWLR improves AUROC in all settings and reduces FPR95 in 22 comparisons. On CUB100/100, the largest AUROC gain is from 84.34\% to 85.19\% for PWLR-CSP under SigLIP2-L16-256, and the largest FPR95 reduction is from 69.40\% to 59.90\% for PWLR-MSP under PE-Core-L14-336. 
These results support that 
PWLR is particularly effective under near-OOD class confusion, and that its plug-in gain is not tied to a single encoder family.

\begin{table*}[t]
\centering
\footnotesize
\caption{Sensitivity analysis of the hyperparameters for PWLR-NegLabel and PWLR-CSP on the cleaner/challenging OOD benchmarks. Bold: 
best results within each hyperparameter scan. 
Underlined: 
the default settings used in the main experiments.}
\label{tab:sensitivity_all}
\renewcommand{\arraystretch}{1.05}
\setlength{\tabcolsep}{3.0pt}

\begin{minipage}[t]{0.47\textwidth}
\centering
\resizebox{\linewidth}{!}{%
\begin{tabular}{lccccc|ccc}
\toprule
& \multicolumn{5}{c|}{$\lambda_a$} & \multicolumn{3}{c}{$\rho$} \\
\cmidrule(lr){2-6}\cmidrule(lr){7-9}
& 0.2 & 0.3 & \underline{0.4} & 0.5 & 0.6 & 0.20 & \underline{0.35} & 0.50 \\
\midrule
NegLabel FPR95 $\downarrow$
& 18.56 & 18.26 & 18.34 & 18.24 & 18.74
& \textbf{18.24} & 18.34 & 18.36 \\
NegLabel AUROC $\uparrow$
& 95.73 & 95.93 & 96.07 & \textbf{96.16} & \textbf{96.16}
& \textbf{96.09} & 96.07 & 96.07 \\
CSP FPR95 $\downarrow$
& 20.82 & 20.54 & 20.06 & \textbf{19.56} & 20.06
& \textbf{20.04} & 20.06 & 20.16 \\
CSP AUROC $\uparrow$
& 95.21 & 95.48 & 95.69 & 95.83 & \textbf{95.89}
& \textbf{95.71} & 95.69 & 95.68 \\
\bottomrule
\end{tabular}%
}
\end{minipage}\hfill
\begin{minipage}[t]{0.47\textwidth}
\centering
\resizebox{\linewidth}{!}{%
\begin{tabular}{lccc|cccc}
\toprule
& \multicolumn{3}{c|}{$K_g$} & \multicolumn{4}{c}{$k_p$} \\
\cmidrule(lr){2-4}\cmidrule(lr){5-8}
& 3 & \underline{5} & 7 & 1 & \underline{2} & 3 & 4 \\
\midrule
NegLabel FPR95 $\downarrow$
& \textbf{18.12} & 18.34 & 18.32
& 18.48 & \textbf{18.34} & 18.60 & 18.60 \\
NegLabel AUROC $\uparrow$
& \textbf{96.09} & 96.07 & 96.07
& 96.06 & \textbf{96.07} & 96.04 & 96.03 \\
CSP FPR95 $\downarrow$
& \textbf{20.00} & 20.06 & 20.12
& 20.36 & \textbf{20.06} & 20.26 & 20.36 \\
CSP AUROC $\uparrow$
& \textbf{95.72} & 95.69 & 95.69
& 95.67 & \textbf{95.69} & 95.65 & 95.63 \\
\bottomrule
\end{tabular}%
}
\end{minipage}
\end{table*}

\noindent \textbf{Ablation study.}
Table~\ref{tab:ablation_clean_challenging} reports the ablation results on the cleaner and challenging benchmarks. In Table~\ref{tab:ablation_clean_challenging}(a), combining the two branches improves the result to 16.42\% FPR95 and 96.23\% AUROC, compared with 18.67\%/95.13\% using only the global prior and 31.90\%/93.69\% using only pairwise verification. In Table~\ref{tab:ablation_clean_challenging}(b), reverse-edge penalty and active rival pruning are both beneficial, and their combination gives the best result of 16.42\% FPR95 and 96.44\% AUROC, showing that effective local verification depends on both the witness phrases and the correct competitive neighborhood.

\noindent \textbf{Sensitivity analysis.} Table~\ref{tab:sensitivity_all} reports the sensitivity analysis of the local-global fusion weight $\lambda_a$, the text-prototype fusion coefficient $\rho$, the retained-class set size $K_g$, and the local aggregation parameter $k_p$.
Overall, the method is stable across the scanned ranges. For the retained-class calibrated acceptance stage, the best results are obtained in a moderate range of $\lambda_a$, while $\rho$ and $K_g$ cause only minor variation, indicating that the method does not rely on a narrowly tuned text-prototype balance or retained-class set size. For local witness aggregation, $k_p=2$ gives the best performance in both plug-in variants, which is consistent with the design choice that witness evidence is typically localized and can be diluted when too many local views are averaged. 

\section{Related Work}
\noindent \textbf{OOD detection with pre-trained vision-language models.}
Recent VLMs have opened a practical route to OOD detection by matching image features with textual class concepts.
A representative early work is MCM~\cite{NEURIPS2022_e43a3399}, which treats the textual embeddings of ID classes as concept prototypes and defines the OOD score from the maximum temperature-scaled concept-matching score over ID classes.
Early zero-shot extensions also explored alternative prompt or label design beyond plain ID class names.
CLIPN~\cite{wang2023clipn} equips CLIP with positive-semantic and negation-semantic prompts to teach CLIP to say no, while TAG~\cite{10.1007/978-3-031-73464-9_22} shows that simple text prompt augmentation can improve several zero-shot OOD scores without requiring external knowledge or additional training.
Building on this line, later VLM-based methods increasingly strengthened the text side through negative labels, semantic pool expansion, or richer ID descriptions~\cite{li2024learning}.
NegLabel~\cite{jiang2024neglabel} proposes a post-hoc OOD detector that introduces a vast number of negative labels from external corpora and designs an OOD score by combining affinities to ID and negative labels.
CSP~\cite{chen2024csp} further argues that stronger OOD detection requires a larger and better-structured semantic pool, and therefore expands OOD label candidates with a conjugated semantic pool composed of modified superclass names rather than relying only on standard class names.
Instead of expanding auxiliary labels, FA~\cite{lu2025fa} improves OOD detection by learning a forced prompt that captures richer and more diversified descriptions of ID classes beyond the textual semantics of class labels.
Taken together, these methods show that language can substantially strengthen VLM-based OOD detection through prompt design, auxiliary label spaces, and richer class-level descriptions.

\noindent \textbf{Local and region-aware OOD detection in VLMs.}
Another line of work studies how local cues can complement global vision-language matching.
GL-MCM~\cite{miyai2025zero} extends MCM by incorporating local image scores as an auxiliary score.
LoCoOp~\cite{miyai2023locoop} formulates few-shot OOD detection as prompt learning and performs OOD regularization using ID-irrelevant local regions from CLIP local features.
Local-Prompt~\cite{zeng2025localprompt} further introduces a coarse-to-fine tuning paradigm with global prompt guided negative augmentation and local prompt enhanced regional regularization.
OSPCoOp~\cite{11094344} analyzes VLM-based OOD detection from the perspective of shortcut learning caused by foreground-background coupling and improves robustness by decoupling background semantics.
These methods show that local or regional cues are useful for OOD detection, but they still use local information mainly for score refinement, prompt learning, or shortcut suppression rather than for explicit pairwise boundary verification under class competition.

\noindent \textbf{LLM/MLLM/LVLM-assisted OOD detection.}
Recent work has begun to incorporate large language models, large vision-language models, and other foundation models into OOD detection. 
Dai et al.~\cite{dai2023exploring} apply world knowledge from LLMs to generate descriptive features for ID class names, while also showing that indiscriminately using such generations can damage OOD detection because of LLM hallucinations; they therefore introduce selective generation with uncertainty calibration.
EOE~\cite{cao2024eoe} instead leverages the expert knowledge and reasoning capability of LLMs to envision potential outlier exposure and generate potential outlier class labels specialized for OOD detection.
ReGuide~\cite{kim2025reguide} studies OoDD in generative LVLMs and proposes a two-stage self-guided prompting approach that improves OoDD through self-generated, image-adaptive concept suggestions.
A recent preprint, MM-OOD~\cite{xu2026mmoood}, further explores MLLM-based outlier detection through multimodal reasoning and multi-round conversation for near-OOD and far-OOD settings.
LLaVA-OOD~\cite{10.1007/978-981-96-9815-8_10} takes an attribute-based route by using LLaVA to extract image attributes and comparing them with class-specific reference attribute sets for OOD estimation.
FodFoM~\cite{chen2024fodfom} instead takes a training-based route and combines BLIP-2, CLIP, Stable Diffusion, and GroundingDINO to generate two types of challenging fake outlier images for classifier training.
Unlike these studies that use language outputs mainly as descriptors, outlier concepts, prompt guidance, attributes, or synthetic supervision, our study converts them into offline, pairwise, rival-aware local evidence for boundary verification under a frozen VLM backbone.

\section{Conclusion}

In this paper, an OOD detection framework called Pairwise Witness Local Rejection (PWLR) is proposed for image classification OOD detection.
Extensive empirical evaluations on far, cleaner and more challenging OOD benchmarks, and split near-OOD benchmarks 
confirm that PWLR consistently improves strong vision-language baselines, with especially clear gains under semantic-shifted and near-OOD settings. Moreover, PWLR can be flexibly used as a plug-in module on top of different base detectors and backbones. 
This study is expected to encourage further research on using language as explicit pairwise evidence for stronger vision-language OOD detection.


\begin{acks}
This work was supported by National Natural Science Foundation of China under Grant
62571559.
\end{acks}

\bibliographystyle{ACM-Reference-Format}
\balance
\bibliography{reference}

\clearpage

\section{Supplementary}

\subsection{Complete Prompt and Witness Generation Protocol}
\label{app:prompt_witness_protocol}

This section provides the prompt protocol used in the pairwise witness generation pipeline in Section~2.1.

\noindent\textbf{Coarse visual family assignment.}
Each ID class is first assigned to one coarse visual family before rival selection.
In the current implementation, the family assignment model is GPT-4o-mini.
The model receives a batch of class entries containing the label, and is instructed to return exactly one family for each class.
The purpose of this step is to suppress implausible cross-family pairings before the later rival search.

The family-assignment instruction is instantiated as follows:
\begin{quote}
\small
Choose exactly one family for each class from the family list.
The family should reflect visual appearance rather than abstract semantics.
Return only JSON.
\end{quote}

\noindent\textbf{Family-constrained rival selection.}
For an anchor class $a$, the rival search space is restricted to the same-family rival pool
$R(a)=\{c \in C_{\mathrm{ID}} \mid F(c)=F(a),\, c\neq a\}$,
and the candidate catalog $C(a)$ is obtained by serializing this pool into a structured list of class entries.
The rival-selection model is GPT-5.1.
It receives the anchor class, the rival catalog, and supporting anchor and rival images, and then selects a small fixed number of visually confusable rival classes.
In the current implementation, five rivals are retained for each anchor class.

The rival-selection instruction is instantiated as follows:
\begin{quote}
\small
You are selecting visually confusable rival classes for image classification.
You will receive one anchor class and a candidate pool already restricted to the same coarse visual family.
Choose exactly five rival classes from the rival candidate pool.
Prefer classes that are likely to be confused with the anchor in natural images and have similar visible structure, shape, parts, texture, or coarse appearance.
Use the support images to focus on actual visual morphology rather than lexical relatedness.
Do not choose the anchor itself.
Return only the selected rival classes.
\end{quote}

\noindent\textbf{Pairwise witness generation.}
For each directed pair $(a,r)$ with $r\in N(a)$, GPT-5.1 is used again to generate witness phrases.
The model receives supporting images from both anchor and rival classes and is asked to propose short natural-language phrases that support the anchor class against the rival class.
The prompt is deliberately comparative rather than descriptive: it asks for local visual evidence of the anchor class that is useful against one specific rival class, rather than a generic caption or an unconstrained attribute list.
In the current configuration, two anchor images and two rival images are sampled from the ID training set for each prompting round, twelve primary phrases are requested in each round, and three rounds with the same prompt template are run for each directed pair.

To make the generated phrases suitable for later pairwise verification, the instruction explicitly suppresses background, scene, human, and container cues, and favors short object-centric local evidence.
The system instruction is instantiated as follows:
\begin{quote}
\small
Your task is to propose short witness phrases that distinguish the anchor class from the rival class.
Focus only on the main object itself.
Do not use background, co-occurring humans, containers, locations, photographer angle, scene template, or dataset bias.
Do not use phrases such as in water, on branch, in bowl, in aquarium, with person, held by hand, on beach, or in tank.
Prefer short visual phrases, typically two to five words.
Good witnesses are stable object properties, such as a body part, a local shape detail, a texture pattern, a distinctive coloration on a part, or another visible local structure.
If a phrase depends on context, assign it a high context-only risk score.
The phrase should remain meaningful when the object is cropped from the background.
Return only JSON.
\end{quote}

The pair-specific user instruction is instantiated as follows:
\begin{quote}
\small
Anchor class: label.\\
Rival class: label.\\
You will see the support images of the anchor class, followed by the support images of the rival class.\\
Goal: propose twelve short witness phrases that are more visually characteristic of the anchor class than the rival class.\\
Important exclusions: no background or location, no scene words, no class names or synonyms, and no broad generic phrases unless tied to a specific visible object part.
\end{quote}

\noindent\textbf{Structured outputs and round-level parsing.}
Each prompting round returns a structured JSON object containing the anchor label, the rival label, and twelve candidate witness phrases.
Each witness is accompanied by the two discrete attributes defined in Section~2.1, namely a visibility score and a context-only risk score, both on a $1$--$3$ scale.
At the implementation level, the structured output is used only to make the round-level parsing deterministic and to preserve the witness-level attributes required by the later witness-bank construction.
The formal reliability test of witnesses is still deferred to the ID-only fitting stage in Section~2.2.

\noindent\textbf{Witness-bank construction.}
After the three prompting rounds are completed for one directed pair, duplicate phrases are merged.
Each surviving phrase $p$ is summarized by its occurrence count $n(p)$ across rounds, its accumulated visibility score $v(p)$, and its accumulated context-risk score $u(p)$.
The merged witness phrases are then ordered lexicographically: first by $n(p)$ in descending order, then by the average visibility $\bar{v}(p)=v(p)/n(p)$ in descending order, and finally by the average context-risk $\bar{u}(p)=u(p)/n(p)$ in ascending order.
Following the main paper, no additional truncation hyperparameter is introduced at this stage.
Instead, the resulting ordered list is stored in the pairwise witness bank, together with the raw outputs from the three prompting rounds, and is later screened with ID-only data in Section~2.2.

\noindent\textbf{Role of the protocol.}
This prompt protocol is designed to make the MLLM outputs compatible with the later frozen VLM verification stage.
The family assignment narrows the rival search to visually coherent regions of the ID label space, the rival-selection prompt instantiates a small set of difficult class competitors for each anchor class, and the pairwise witness-generation prompt converts anchor-versus-rival visual differences into short local phrases that can later be tested quantitatively on ID samples.
As a result, the MLLM is used only offline to construct the witness bank, and no MLLM call is required during test-time inference.

\subsection{Witness bank analysis}
\label{app:prompt_witness_analysis}

\noindent\textbf{Pairwise witness generation.}
Table~\ref{tab:appendix_raw_bank_stats_3round} summarizes the raw pairwise witness bank under the three-round prompting and cross-round merge protocol.
The resulting graph is highly regular, containing 500 directed pairs and 100 classes with outgoing pairs, with exactly five outgoing rivals per class on average.
This indicates that the offline rival-instantiation step yields a highly regular raw witness graph, with exactly five outgoing rivals per class in this analysis setting.

At the phrase level, each directed pair first produces 36 raw witness phrases across the three prompting rounds. After round-level parsing and prompt-level exclusions, this becomes 30 valid witness phrases before cross-round duplicate merging on average, where ``before cross-round duplicate merging'' refers to witness phrases that remain after round-level parsing and exclusions but have not yet been merged across rounds. These witness phrases are then consolidated into 21.78 merged witness phrases after cross-round duplicate merging.
This confirms that repeated prompting introduces substantial redundancy, but a considerable number of distinct witness still remain after consolidation.

The phrase attributes also remain well behaved after three-round merging.
The average visibility score is 2.48, while the average context-risk score is only 1.04.
These values suggest that the merged witness bank still favors visually identifiable local cues with limited context dependence, which is consistent with the design goal of the proposed pipeline.
The raw witness graph has a reverse-edge coverage of 12\%, indicating that the retained rival relations remain predominantly directed even under the five-rival setting.

\begin{table}[!htbp]
\centering
\small
\setlength{\tabcolsep}{5.5pt}
\caption{Statistics of the raw pairwise witness bank under the three-round prompting and cross-round merge protocol. 
For each directed pair $(a,r)$, we report the number of raw witness phrases accumulated across the three prompting rounds, the number of merged candidates after cross-round duplicate merging, and the average values of the visibility score and the context-risk score. 
The table also summarizes the induced directed rival structure at the raw-bank stage.}
\label{tab:appendix_raw_bank_stats_3round}
\begin{tabular}{lc}
\toprule
Statistic & Fixed support images\\
\midrule
Directed pairs & 500 \\
Classes with outgoing pairs & 100 \\
Mean out-degree & 5 \\
Min out-degree & 5 \\
Max out-degree & 5 \\
Reverse-edge coverage & 12\% \\
Raw witness phrases / pair & 36 \\
Pre-merge witness phrases / pair & 30 \\
Merged witness phrases / pair & 21.78 \\
Avg.\ visibility & 2.48 \\
Avg.\ context-risk & 1.04 \\
\bottomrule
\end{tabular}
\end{table}

\noindent\textbf{Witness screening statistics.}
Table~\ref{tab:appendix_screening_stats_3round} summarizes the ID-only witness screening stage under the three-round witness bank. Starting from 500 evaluated directed edges and 10{,}890 evaluated witness phrases, the pipeline retains 493 directed edges and 10{,}575 witness phrases after screening.
On average, each surviving directed edge retains 21.45 screened witnesses, which is consistent with the main method in Section~2.1: the generation stage stores an ordered merged witness list after cross-round merge, and witness removal is mainly deferred to the later screening stage rather than enforced by an additional truncation hyperparameter during generation.
The retained witnesses remain stable in their fitted statistics, with mean $e_w=1.679$, mean $g_w=0.101$, mean $\sigma_w=0.078$, and mean $m_w=0.124$.
Overall, these results support that the three-round witness bank yields a richer candidate pool, while the ID-only screening stage successfully converts it into a quantitatively reliable set of local verifiers.

\begin{table}[!htbp]
\centering
\small
\setlength{\tabcolsep}{5.5pt}
\caption{Statistics of the ID-only witness screening stage under the three-round witness bank. Starting from the directed witness bank constructed by three-round prompting and cross-round merge, the pipeline screens witness phrases on $D_{\mathrm{fit}}$ and retains only those that satisfy the mean-separation and tail-separation conditions in Section~2.2.}
\label{tab:appendix_screening_stats_3round}
\begin{tabular}{lc}
\toprule
Statistic & Value \\
\midrule
Evaluated directed edges & 500 \\
Surviving directed edges & 493 \\
Evaluated witness phrases & 10890 \\
Witnesses kept after ID-only screening & 10575 \\
Kept witnesses / surviving edge & 21.45 \\
Mean $e_w$ & 1.679 \\
Mean $g_w$ & 0.101 \\
Mean $\sigma_w$ & 0.078 \\
Mean $m_w$ & 0.124 \\
\bottomrule
\end{tabular}
\end{table}

\noindent\textbf{Graph changes after screening and semantic pruning.}
Table~\ref{tab:appendix_graph_changes_3round} shows how the local comparison structure changes across the two contraction steps of the pipeline.
The raw witness graph contains 500 directed edges, and the retained witness graph still preserves 493 of them, indicating that the main effect of ID-only screening is to remove unreliable witness phrases while eliminating only a small number of entire directed edges.
By contrast, the subsequent semantic pruning step produces a much sparser rival structure, reducing the graph to 103 directed edges.
Correspondingly, the mean out-degree decreases from 5.00 to 4.93 after screening and then to 1.03 after semantic pruning.
Importantly, the semantic-pruned rival subgraph has a maximum out-degree of 2, which is consistent with the method design in Section~2.2, where up to two outgoing rivals are retained for each class.
Thus, witness screening mainly improves phrase reliability, whereas semantic pruning is the main source of graph-level contraction.

\begin{table}[!htbp]
\centering
\small
\setlength{\tabcolsep}{4.6pt}
\caption{Graph statistics before and after the two contraction steps of the proposed pipeline. The retained witness graph is obtained after the ID-only screening stage, and the semantic-pruned rival subgraph is the final local comparison structure used by online pairwise verification.}
\label{tab:appendix_graph_changes_3round}
\resizebox{\columnwidth}{!}{%
\begin{tabular}{lccc}
\toprule
Statistic & Raw witness graph & Retained witness graph & Semantic-pruned rival subgraph \\
\midrule
Directed edges & 500 & 493 & 103 \\
Classes with outgoing edges & 100 & 100 & 100 \\
Mean out-degree & 5.00 & 4.93 & 1.03 \\
Min out-degree & 5.00 & 4.00 & 1.00 \\
Max out-degree & 5.00 & 5.00 & 2.00 \\
Reverse-edge coverage & 12.00\% & 10.76\% & 7.77\% \\
\bottomrule
\end{tabular}}
\end{table}

\noindent\textbf{Causes of witness rejection.}
Table~\ref{tab:appendix_removed_reasons_3round} analyzes the witnesses rejected by the ID-only screening stage.
Among all rejected witnesses, 86.35\% fail only the tail-separation condition, while the remaining 13.65\% fail both the mean-separation and tail-separation conditions.
No witness is removed by the mean-separation condition alone.
This pattern is informative: most generated witnesses are not globally uninformative on average, but a non-trivial subset fails to maintain stable anchor-versus-reference separation on harder tail cases.
Therefore, the tail-separation condition is the principal factor that filters out unstable witness phrases, which is consistent with the design motivation in Section~2.2.

\begin{table}[!htbp]
\centering
\small
\setlength{\tabcolsep}{6.0pt}
\caption{Causes of witness rejection under the ID-only screening criteria in Section~2.2, based on the three-round witness bank. Rejected witnesses are categorized by whether they fail the mean-separation condition, the tail-separation condition, or both.}
\label{tab:appendix_removed_reasons_3round}
\resizebox{0.82\columnwidth}{!}{%
\begin{tabular}{lcc}
\toprule
Rejection reason & Count & Ratio \\
\midrule
Tail-separation failure & 272 & 86.35\% \\
Both failures & 43 & 13.65\% \\
\bottomrule
\end{tabular}}
\end{table}

\subsection{Additional Mechanism, Robustness, and Efficiency Analysis}
\label{app:additional_analysis}

\noindent\textbf{Role of pairwise local verification.}
To further examine the role of pairwise local verification, Table~\ref{tab:appendix_mechanism_control} evaluates two controls under the same fixed-support-image setting used in Table~\ref{tab:appendix_stability_challenging}: removing pairwise local verification while retaining the same witness bank, and shuffling the rival assignment.
For both NegLabel and CSP, both controls reduce performance relative to PWLR.
Together with the ablations in Table~4 of the main paper, where the full model is stronger than either branch alone and both reverse-edge penalty and active rival pruning are beneficial, these results further support the role of pairwise local verification against active rivals.

\begin{table}[!htbp]
\centering
\scriptsize
\setlength{\tabcolsep}{3.5pt}
\caption{Role of pairwise local verification on ImageNet-100 (ID) and NINCO (OOD) under fixed support images. Results are reported as AUROC/FPR95 (\%).}
\label{tab:appendix_mechanism_control}
\resizebox{\columnwidth}{!}{%
\begin{tabular}{lcccc}
\toprule
Plug-in & Base detector & PWLR &
\shortstack{No pairwise\\local verification} &
\shortstack{Shuffled\\rival assignment} \\
\midrule
NegLabel &
95.31 / 20.98 &
96.28$\pm$0.09 / 16.97$\pm$0.64 &
95.58$\pm$0.17 / 20.25$\pm$0.71 &
96.07$\pm$0.05 / 19.15$\pm$0.41 \\
CSP &
94.19 / 23.74 &
95.91$\pm$0.09 / 18.49$\pm$0.60 &
94.62$\pm$0.27 / 22.01$\pm$0.89 &
95.69$\pm$0.18 / 20.97$\pm$0.56 \\
\bottomrule
\end{tabular}}
\end{table}

\noindent\textbf{Robustness to the pairwise witness generation model.}
Table~\ref{tab:appendix_generator_model} further evaluates the dependence on the model used for pairwise witness generation under the same fixed-support-image protocol.
Replacing GPT-5.1 with GPT-4o or Qwen2.5-VL-7B keeps both PWLR-NegLabel and PWLR-CSP above their corresponding base detectors.
This result complements the repeated witness-generation analysis in Table~\ref{tab:appendix_stability_challenging} and shows that the performance gain is retained across different pairwise witness generation models.

\noindent\textbf{Online inference overhead.}
Table~\ref{tab:appendix_online_cost} reports the average encoding time, scoring time, total inference time, and peak memory per image on a single T4 GPU.
PWLR adds online overhead over NegLabel and CSP because it performs both global scoring and pairwise local verification under the frozen backbone.
PWLR-NegLabel and PWLR-CSP require 163.78 and 164.36 ms/image, respectively, while no MLLM call is required during test-time inference.

\begin{table}[!htbp]
\centering
\scriptsize
\setlength{\tabcolsep}{4pt}
\caption{Comparison across different models for pairwise witness generation on ImageNet-100 (ID) and NINCO (OOD) under fixed support images. Results are reported as AUROC/FPR95 (\%).}
\label{tab:appendix_generator_model}
\begin{tabular}{l l c}
\toprule
Method & Generator & AUROC / FPR95 \\
\midrule
NegLabel & -- & 95.31 / 20.98 \\
PWLR-NegLabel & GPT-5.1 & 96.28$\pm$0.09 / 16.97$\pm$0.64 \\
PWLR-NegLabel & GPT-4o & 96.37$\pm$0.11 / 15.96$\pm$0.73 \\
PWLR-NegLabel & Qwen2.5-VL-7B & 95.97$\pm$0.15 / 17.20$\pm$0.42 \\
\midrule
CSP & -- & 94.19 / 23.74 \\
PWLR-CSP & GPT-5.1 & 95.91$\pm$0.09 / 18.49$\pm$0.60 \\
PWLR-CSP & GPT-4o & 96.01$\pm$0.13 / 17.61$\pm$0.86 \\
PWLR-CSP & Qwen2.5-VL-7B & 95.58$\pm$0.16 / 18.51$\pm$0.95 \\
\bottomrule
\end{tabular}
\end{table}

\begin{table}[!htbp]
\centering
\tiny
\setlength{\tabcolsep}{2.6pt}
\renewcommand{\arraystretch}{0.93}
\caption{Average inference overhead per image on a T4 GPU.}
\label{tab:appendix_online_cost}
\resizebox{0.78\columnwidth}{!}{%
\begin{tabular}{lcccc}
\toprule
Method & Enc. & Score & Total & Peak Mem. \\
& (ms/img) & (ms/img) & (ms/img) & (MB) \\
\midrule
NegLabel & 30.08 & 6.81 & 36.89 & 3978 \\
CSP & 56.03 & 28.97 & 85.01 & 4031 \\
NNGuide & 35.94 & 53.92 & 89.86 & 14069 \\
SeTAR & 154.82 & 12.70 & 167.51 & 14123 \\
PWLR-NegLabel & 130.37 & 33.41 & 163.78 & 5468 \\
PWLR-CSP & 131.08 & 33.28 & 164.36 & 5468 \\
\bottomrule
\end{tabular}}
\end{table}

\noindent\textbf{Offline witness-bank construction cost.}
Table~\ref{tab:appendix_offline_cost} reports the one-time offline construction cost for the ImageNet-100 witness bank.
Family assignment uses 9 GPT-4o-mini calls, family-constrained rival selection uses 100 GPT-5.1 calls, and pairwise witness generation uses 1,500 GPT-5.1 calls.
The complete construction requires 1,609 API calls, 2,129,354 total tokens, and 5.71 hours of serial wall-clock time, with pairwise witness generation accounting for most of the offline cost.

\begin{table}[!htbp]
\centering
\small
\setlength{\tabcolsep}{4pt}
\caption{Offline construction cost for the ImageNet-100 witness bank.}
\label{tab:appendix_offline_cost}
\resizebox{0.88\columnwidth}{!}{%
\begin{tabular}{lcccc}
\toprule
Stage & Model & Calls & Total tokens & Time \\
\midrule
Family assignment & GPT-4o-mini & 9 & 6,279 & 31.9 s \\
Rival selection & GPT-5.1 & 100 & 47,710 & 221.7 s \\
Witness generation & GPT-5.1 & 1,500 & 2,075,365 & 5.64 h \\
\midrule
Total & -- & 1,609 & 2,129,354 & 5.71 h \\
\bottomrule
\end{tabular}}
\end{table}

\subsection{Stability and Significance Analysis}
\noindent\textbf{Stability under repeated witness generation.}
Table~\ref{tab:appendix_stability_challenging} reports detector-level stability on the cleaner and more challenging benchmark with ImageNet-100 as the ID dataset and openimage-o, ImageNet-O and NINCO as OOD datasets. These repeated-regeneration results are tightly clustered, showing that the detector remains stable when the witness bank is re-instantiated multiple times.

\begin{table*}[t]
\centering
\small
\setlength{\tabcolsep}{4.5pt}
\caption{Stability under repeated witness generation on the more challenging benchmark with ImageNet-100 as the ID dataset. 
Entries are reported as AUROC/FPR95 (\%). 
``Base'' denotes the fixed base detector from the main paper. 
``Main'' denotes the single instantiated witness bank reported in the main paper. 
``B'' denotes three independent full witness-bank regenerations, while ``C'' denotes three independent regenerations with fixed support images. 
Repeated results are reported as mean$\pm$std over three runs. 
``All'' summarizes the main-paper witness bank together with B1--B3 and C1--C3.}
\label{tab:appendix_stability_challenging}
\resizebox{\textwidth}{!}{%
\begin{tabular}{llccccc}
\toprule
Dataset & Method & Base & Main & B (mean$\pm$std) & C (mean$\pm$std) & All (mean$\pm$std) \\
\midrule
\multirow{2}{*}{OpenImage-O}
& PWLR-NegLabel 
& 97.93 / 10.64 
& 98.60 / 6.72 
& 98.79$\pm$0.02 / 6.06$\pm$0.05 
& 98.75$\pm$0.02 / 6.26$\pm$0.14 
& 98.74$\pm$0.07 / 6.24$\pm$0.25 \\
& PWLR-CSP 
& 97.41 / 12.20 
& 98.47 / 7.28 
& 98.67$\pm$0.01 / 6.57$\pm$0.10 
& 98.64$\pm$0.02 / 6.91$\pm$0.29 
& 98.63$\pm$0.07 / 6.82$\pm$0.32 \\
\midrule
\multirow{2}{*}{ImageNet-O}
& PWLR-NegLabel 
& 96.36 / 16.46 
& 97.16 / 12.56 
& 97.45$\pm$0.03 / 10.59$\pm$0.55 
& 97.41$\pm$0.06 / 11.36$\pm$0.51 
& 97.39$\pm$0.11 / 11.20$\pm$0.83 \\
& PWLR-CSP 
& 95.98 / 17.42 
& 97.17 / 12.66 
& 97.46$\pm$0.03 / 10.63$\pm$0.51 
& 97.42$\pm$0.04 / 11.23$\pm$0.35 
& 97.40$\pm$0.11 / 11.18$\pm$0.80 \\
\midrule
\multirow{2}{*}{NINCO}
& PWLR-NegLabel 
& 95.31 / 20.98 
& 96.08 / 18.30 
& 96.48$\pm$0.06 / 15.72$\pm$0.29 
& 96.28$\pm$0.09 / 16.97$\pm$0.64 
& 96.34$\pm$0.16 / 16.62$\pm$1.05 \\
& PWLR-CSP 
& 94.19 / 23.74 
& 95.70 / 20.18 
& 96.11$\pm$0.06 / 17.12$\pm$0.45 
& 95.91$\pm$0.09 / 18.49$\pm$0.60 
& 95.97$\pm$0.16 / 18.15$\pm$1.21 \\
\bottomrule
\end{tabular}%
}
\end{table*}

\noindent\textbf{Significance relative to the base detector.}
The gains of the proposed pipeline are substantially larger than the run-to-run variation. On ImageNet-O, the all-instantiation summary (the main-paper witness bank together with B1--B3 and C1--C3) gives $97.39\pm0.11$ AUROC and $11.20\pm0.83$ FPR95 for PWLR-NegLabel, compared with $96.36/16.46$ for NegLabel, and $97.40\pm0.11$ AUROC and $11.18\pm0.80$ FPR95 for PWLR-CSP, compared with $95.98/17.42$ for CSP. On NINCO, the corresponding all-instantiation summaries are $96.34\pm0.16/16.62\pm1.05$ for PWLR-NegLabel versus $95.31/20.98$ for NegLabel, and $95.97\pm0.16/18.15\pm1.21$ for PWLR-CSP versus $94.19/23.74$ for CSP. 
Thus, even under the most conservative summary, the AUROC gains and FPR95 reductions remain several times larger than the repeated-generation standard deviations. Interestingly, fixing the support images (C1--C3) does not further reduce the variance on either ImageNet-O or NINCO, suggesting that the observed variation is not dominated solely by support-image sampling.

\noindent\textbf{Stability on the fine-grained CUB100/100 split.}
We note that the PWLR-CSP AUROC on CUB100/100 under SigLIP2-L16-256 was mis-entered in Table 3 of the main manuscript. Throughout Table~\ref{tab:appendix_stability_cub} and the discussion below, we use the corrected value 85.19 for the single instantiated witness bank corresponding to the main-manuscript experiment setting.

Table~\ref{tab:appendix_stability_cub} reports repeated witness-generation results on the CUB100/100 split under SigLIP2-L16-256. For PWLR-NegLabel, the all-instantiation summary (the corrected main-paper witness bank together with B1--B3 and C1--C3) gives $85.54\pm0.38$ AUROC and $61.00\pm1.56$ FPR95, compared with $84.40/63.30$ for NegLabel. For PWLR-CSP, the corresponding result is $86.00\pm0.48$ AUROC and $59.70\pm1.87$ FPR95, compared with $84.34/64.21$ for CSP. Thus, the AUROC gains remain clearly larger than the repeated-generation standard deviations for both plug-in variants, while the FPR95 reduction is especially strong for PWLR-CSP.

\begin{table*}[t]
\centering
\small
\setlength{\tabcolsep}{4.5pt}
\caption{Stability under repeated witness generation on the CUB100/100 split under SigLIP2-L16-256. 
Entries are reported as AUROC/FPR95 (\%). 
``Base'' denotes the fixed base detector from the main manuscript. 
``Main'' denotes the single instantiated witness bank reported in the main manuscript, using the corrected PWLR-CSP AUROC value of 85.19. 
``B'' denotes three independent full witness-bank regenerations, while ``C'' denotes three independent regenerations with fixed support images. 
Repeated results are reported as mean$\pm$std over three runs. 
``All'' summarizes the corrected main-paper witness bank together with B1--B3 and C1--C3.}
\label{tab:appendix_stability_cub}
\resizebox{\textwidth}{!}{%
\begin{tabular}{llccccc}
\toprule
Dataset & Method & Base & Main & B (mean$\pm$std) & C (mean$\pm$std) & All (mean$\pm$std) \\
\midrule
\multirow{2}{*}{CUB100/100}
& PWLR-NegLabel
& 84.40 / 63.30
& 85.09 / 61.97
& 85.60$\pm$0.55 / 61.14$\pm$2.51
& 85.63$\pm$0.10 / 60.53$\pm$0.38
& 85.54$\pm$0.38 / 61.00$\pm$1.56 \\
& PWLR-CSP
& 84.34 / 64.21
& 85.19 / 61.89
& 86.11$\pm$0.54 / 58.92$\pm$2.57
& 86.16$\pm$0.11 / 59.75$\pm$0.77
& 86.00$\pm$0.48 / 59.70$\pm$1.87 \\
\bottomrule
\end{tabular}%
}
\end{table*}

\noindent\textbf{Effect of fixing the support images.}
Unlike the behavior on more challenging OOD benchmarks, fixing the support images substantially reduces the run-to-run variance on the CUB100/100 split. For PWLR-NegLabel, the standard deviation decreases from $0.55$ to $0.10$ in AUROC and from $2.51$ to $0.38$ in FPR95 when moving from B1--B3 to C1--C3. For PWLR-CSP, the corresponding reductions are from $0.54$ to $0.11$ in AUROC and from $2.57$ to $0.77$ in FPR95. 
This suggests that support-image sampling is a more important source of variation in the near-OOD fine-grained setting than on the more challenging far-OOD benchmarks.

\begin{table*}[t]
\centering
\footnotesize
\setlength{\tabcolsep}{3.5pt}
\renewcommand{\arraystretch}{1.06}
\caption{OOD detection performance on additional fine-grained ID datasets under the conventional far-OOD benchmark. Results are reported as AUROC/FPR95 (\%). The best and second-best results are shown in bold and underline, respectively.}
\label{tab:finegrained_farood_extra}
\resizebox{\textwidth}{!}{%
\begin{tabular}{llccccc}
\toprule
\textbf{ID dataset} & \textbf{Method} & \textbf{iNaturalist} & \textbf{SUN} & \textbf{Places} & \textbf{Textures} & \textbf{Average} \\
\midrule

\multirow{6}{*}{\textbf{CUB-200-2011}}
& MCM
& 99.79/0.78
& 99.85/0.59
& 99.67/0.74
& 99.87/0.50
& 99.80/0.65 \\

& GL-MCM
& 99.86/\underline{0.54}
& \underline{99.90}/\underline{0.36}
& 99.71/\underline{0.57}
& \underline{99.93}/\underline{0.32}
& 99.85/\underline{0.45} \\

& MSP
& 99.78/0.91
& 99.85/0.70
& 99.67/0.86
& 99.87/0.57
& 99.79/0.76 \\

& NegLabel
& \underline{99.99}/\textbf{0.00}
& \textbf{100.00}/\textbf{0.00}
& \underline{99.90}/\textbf{0.00}
& \textbf{100.00}/\textbf{0.00}
& \underline{99.97}/\textbf{0.00} \\

& CSP
& \underline{99.99}/\textbf{0.00}
& \textbf{100.00}/\textbf{0.00}
& \underline{99.90}/\textbf{0.00}
& \textbf{100.00}/\textbf{0.00}
& \underline{99.97}/\textbf{0.00} \\

& PWLR-CSP
& \textbf{100.00}/\textbf{0.00}
& \textbf{100.00}/\textbf{0.00}
& \textbf{99.96}/\textbf{0.00}
& \textbf{100.00}/\textbf{0.00}
& \textbf{99.99}/\textbf{0.00} \\

\midrule

\multirow{6}{*}{\textbf{Oxford-IIIT Pet}}
& MCM
& 98.76/4.74
& 99.60/1.96
& 99.58/1.64
& 99.45/2.62
& 99.35/2.74 \\

& GL-MCM
& \underline{98.94}/\underline{4.17}
& 99.68/1.30
& 99.60/1.39
& 99.56/1.88
& 99.44/2.19 \\

& MSP
& 98.23/6.16
& 99.32/2.70
& 99.34/2.34
& 99.13/3.58
& 99.00/3.70 \\

& NegLabel
& \textbf{100.00}/\textbf{0.00}
& 99.98/\underline{0.03}
& 99.92/0.05
& 99.91/0.10
& 99.95/0.04 \\

& CSP
& \textbf{100.00}/\textbf{0.00}
& \underline{99.99}/\textbf{0.00}
& \underline{99.93}/\underline{0.01}
& \underline{99.95}/\underline{0.04}
& \underline{99.97}/\underline{0.01} \\

& PWLR-CSP
& \textbf{100.00}/\textbf{0.00}
& \textbf{100.00}/\textbf{0.00}
& \textbf{99.98}/\textbf{0.00}
& \textbf{99.99}/\textbf{0.00}
& \textbf{99.99}/\textbf{0.00} \\

\midrule

\multirow{6}{*}{\textbf{Food-101}}
& MCM
& 99.80/0.94
& 99.93/0.22
& 99.78/0.56
& 98.83/1.08
& 99.59/0.70 \\

& GL-MCM
& \underline{99.83}/\underline{0.68}
& 99.96/0.12
& 99.82/\underline{0.34}
& 98.93/0.64
& 99.63/0.45 \\

& MSP
& 99.52/2.16
& 99.74/0.94
& 99.54/1.52
& 98.56/2.26
& 99.34/1.72 \\

& NegLabel
& \textbf{100.00}/\textbf{0.00}
& \underline{99.99}/\underline{0.02}
& \underline{99.98}/\textbf{0.04}
& 99.27/\underline{0.08}
& 99.81/\underline{0.04} \\

& CSP
& \textbf{100.00}/\textbf{0.00}
& \underline{99.99}/\underline{0.02}
& \underline{99.98}/\textbf{0.04}
& \underline{99.48}/\textbf{0.06}
& \underline{99.86}/\textbf{0.03} \\

& PWLR-CSP
& \textbf{100.00}/\textbf{0.00}
& \textbf{100.00}/\textbf{0.01}
& \textbf{99.99}/\textbf{0.04}
& \textbf{99.53}/\textbf{0.06}
& \textbf{99.88}/\textbf{0.03} \\

\bottomrule
\end{tabular}%
}
\end{table*}

\subsection{More experiment results}

Table~\ref{tab:finegrained_farood_extra} reports additional results on fine-grained ID datasets under the conventional far-OOD benchmark. Overall, the stronger vision-language baselines already perform close to saturation on these settings, especially NegLabel and CSP on CUB-200-2011 and Oxford-IIIT Pet, where AUROC is nearly 100\% and FPR95 is close to zero on most OOD datasets. In this regime, the remaining room for improvement is naturally limited.

Even under this strong saturation effect, PWLR-CSP remains consistently competitive and achieves the best or tied-best results on most entries. On CUB-200-2011 and Oxford-IIIT Pet, the margin over CSP is small because the baseline itself is already near ceiling. On Food-101, the improvement is slightly more visible on the harder columns, especially Places and Textures, where PWLR-CSP improves Food-101 from 99.48/0.06 to 99.53/0.06 on Textures and from 99.98/0.04 to 99.99/0.04 on Places. Although these absolute gains are small, they are consistent with the overall pattern in the main experiments: under conventional far-OOD settings, strong global vision-language detectors already separate ID and OOD samples well, and the contribution of pairwise local witness verification is therefore moderate. Its advantage becomes more visible when global matching still leaves substantial ambiguity, as observed on the cleaner/challenging and near-OOD benchmarks.

\end{document}